%% file: latex/acl_latex.tex
\pdfoutput=1

\documentclass[11pt]{article}

\usepackage[final]{acl}

\usepackage{times}
\usepackage{latexsym}

\usepackage[T1]{fontenc}

\usepackage[utf8]{inputenc}

\usepackage{microtype}

\usepackage{inconsolata}

\usepackage[utf8]{inputenc}
\usepackage{CJKutf8}

\usepackage{enumitem}
\usepackage{caption}
\usepackage{subcaption}
\usepackage{graphicx}
\usepackage[inkscapelatex=false]{svg}
\usepackage{booktabs}
\usepackage{multirow}
\usepackage{array}
\usepackage{amsmath}
\usepackage{arydshln} 

\definecolor{deepgreen}{rgb}{0.1,0.85,0.1}
\definecolor{shun}{rgb}{0.1,0.8,0.1}

\usepackage{stfloats}

\usepackage{pifont}
\usepackage{makecell}

\title{Exploring Task Performance with Interpretable Models via Sparse Auto-Encoders}

\author{Shun Wang\textsuperscript{1},  Tyler Loakman\textsuperscript{1}, Youbo Lei\textsuperscript{2},  Yi Liu\textsuperscript{3}, Bohao Yang\textsuperscript{4}, Yuting Zhao\textsuperscript{5}, \\
Dong Yang\textsuperscript{6$*$}, Chenghua Lin\textsuperscript{1,4}\thanks{Corresponding author}\\
\textsuperscript{1}Department of Computer Science,
The University of Sheffield, UK \\
\textsuperscript{2}Xi'an Jiaotong University  ~~~ 
\textsuperscript{3}Tianjin University ~~~ 
\textsuperscript{4}University of Manchester
\\
\textsuperscript{5}Department of Advanced Information Technology, Kyushu University\\
\textsuperscript{6}City University of Hong Kong\\
\texttt{swang209@sheffield.ac.uk}, ~~~ \texttt{chenghua.lin@manchester.ac.uk}
}
\begin{document}
{\makeatletter\acl@anonymizefalse
  \maketitle
}

\input{latex/0_abs}
\input{latex/1_intro}

\input{latex/2_related_works}
\input{latex/3_methodology}
\input{latex/4_experiments}

\input{latex/5_results}

\input{latex/6_conclusions}

\section*{Limitations}
In this work, we apply pre-existing interpretability techniques on real LLMs, moving away from the toy models that have been used in existing works. However, as a result, we do not present a new interpretability technique, but rather a novel framework for applying these techniques to real-world downstream tasks in order to quantify how they can be used to improve existing models through the identification of model understanding. 
Furthermore, we test this approach on two specific downstream tasks, mathematical reasoning and metaphor detection, due to representing logical reasoning and abstract figurative language understanding. However, it is possible that varying performance impacts could be found for other tasks that we have not investigated in the present work.

\section*{Ethics Statement}
We believe in and firmly adhere to the Code of Conduct in the performance of this work and the methods involved.


\bibliography{custom}

\clearpage
\appendix
\input{latex/a_appendix}

\end{document}

%% file: latex/0_abs.tex
\begin{abstract}
Large Language Models (LLMs) are traditionally viewed as black-box algorithms, therefore reducing trustworthiness and obscuring potential approaches to increasing performance on downstream tasks. In this work, we apply an effective LLM decomposition method using a dictionary-learning approach with sparse autoencoders. This helps extract monosemantic features from polysemantic LLM neurons. Remarkably, our work identifies model-internal misunderstanding, allowing the automatic reformulation of the prompts with additional annotations to improve the interpretation by LLMs. Moreover, this approach demonstrates a significant performance improvement in downstream tasks, such as mathematical reasoning and metaphor detection.
\end{abstract}

%% file: latex/1_intro.tex
\section{Introduction}
The interpretability of Large Language Models (LLMs) is a critical area of research. Despite their remarkable performance across a range of tasks, these models remain inherently opaque \cite{li2024bladeenhancingblackboxlarge}. The ability to understand the internal decision-making processes and representational structures of LLMs is essential for building trust, ensuring safety, and optimizing performance, particularly in sensitive fields such as healthcare~\cite{nazi2024large} and law~\cite{shu2024lawllm}. 
This requirement for interpretability stems from the black-box nature of neural networks, where the sheer complexity 
poses a significant challenge in understanding how input data is processed and represented within the model. Addressing this opacity is a key focus for advancing the responsible deployment of LLMs in real-world applications~\cite{lin2024towards}.

LLMs such as GPT-4 \cite{OpenAI_GPT4_2023} and Llama 3 \cite{dubey2024llama3herdmodels} are trained on massive datasets to capture intricate language patterns. However, their opacity (i.e., their inability to be easily understood by humans) remains a significant limitation. \citet{rudin2019stop} argues against using black-box models in sensitive fields, advocating for inherently interpretable models that do not require post-hoc explanations which may not meet legal criteria for transparency \cite{post-hoc-legal}. Similarly, \citet{lipton2018mythos} critiques the over-reliance on post-hoc interpretability techniques. These concerns have driven research on new methods for breaking down the internal mechanics of LLMs to understand how they process information.

OpenAI proposed a method for using large models to interpret other large models~\cite{bills2023language}. They utilized GPT-4 to analyze the internal structure of GPT-2 to reveal more information. However, as noted in their report, this simplistic approach primarily relies on analyzing how individual neurons respond to specific features in the data, which presents certain limitations. As discussed in~\citet{elhage2022toy}, neurons are often polysemantic, meaning they respond to multiple unrelated inputs, making it difficult to interpret their behavior based on individual activations alone~\cite{bricken2023towards}. Since each neuron is typically activated by multiple features, it is challenging to determine the precise proportion of activations attributed to each feature and their impact on the output. Furthermore, this limitation hinders the application of interpretability methods to improve downstream tasks, as the effect of activating a particular node is multifaceted, forming a one-to-many mapping. Consequently, while it is possible to infer the behavior of neurons from the output retrospectively, it remains difficult to predict the output state directly from the activation of individual neurons.

To address this issue, inspired by the work of~\citeauthor{bricken2023towards}, we apply a mechanistic interpretability technique that combines dictionary learning with sparse autoencoders to decompose polysemantic neurons in various LLMs. One possible explanation for the polysemantic nature of neurons is superposition \cite{arora2018linear, olah2020zoom, elhage2022toy, Xiong_everything_everywhere}, where a neural network represents more independent features than it has neurons by assigning each feature a unique linear combination of neurons. In this scenario, features can be considered vectors spanning multiple neurons, forming an overcomplete basis for activations. Prior studies on toy models \cite{elhage2022toy} demonstrated that when features in the data are sparse, the training process may encourage the model to utilize superposition, facilitating the disambiguation of feature combinations responsible for specific activations.

We integrate dictionary learning with automated interpretation techniques to address the critical need for interpretability in LLMs, explicitly decomposing and analyzing model features. Specifically, we train sparse autoencoders to extract monosemantic features from polysemantic neurons, thereby reducing the ambiguity in individual neuron behavior. Additionally, we provide a comprehensive unsupervised framework for applying mechanistic interpretability analyses to downstream tasks, enabling a clearer understanding of these models' internal mechanisms.

Our contributions are distinct from prior work in several key aspects:
\begin{itemize}

    \item \textbf{Decomposition of Advanced LLMs}: Unlike existing studies focussing on toy models, we decompose several advanced open-source LLMs, directly comparing their performance on downstream tasks.
    
    \item \textbf{Modular Application to Downstream Tasks}: We are the first to apply autoencoder-based feature decomposition in a modular manner to complex downstream tasks, demonstrating improved performance.


    \item \textbf{Ambiguity Resolution in Mathematical Symbols}: By decomposing LLMs, we discovered that they exhibit ambiguities when processing certain mathematical symbols, hindering their reasoning abilities. We resolve these ambiguities through query reformulation, improving the performance of these LLMs in mathematical problem-solving.

    \item \textbf{Enhanced Metaphor Detection}: Our method leverages internal feature extraction rather than external tools like POS tagging or knowledge graphs~\cite{tian2024theory}, providing a more direct and accurate analysis of how LLMs understand metaphors. 
\end{itemize}
Overall, our work introduces a novel generalisable framework that advances the interpretability of LLMs while simultaneously enhancing their performance in critical downstream tasks.

%% file: latex/2_related_works.tex
\section{Related Work}

\noindent\textbf{Interpretability of LLMs}~~\citet{belinkov2019analysis} reviewed interpretability methods in neural language models, highlighting the need to disentangle features for a clearer understanding of LLMs' internal structures. Techniques like attention visualization, layer-wise relevance propagation, and probing classifiers provide partial insights but fail to fully explain model decision-making~\cite{karpathy2015visualizing, gupta2018lisa, conneau2018you}. In contrast, sparse representations and feature decomposition show greater potential for producing human-readable interpretations~\cite{faruqui2015sparse, murdoch2018beyond}. A key challenge in mechanistic interpretability is polysemanticity, where neurons respond to multiple unrelated inputs, complicating function assignment~\cite{olah2020zoom}. \citeauthor{olah2020zoom} introduced the concept of circuits in LLMs, demonstrating how analyzing neuron activations and their interactions can clarify model behavior.



Prior research suggests that polysemantic neurons arise from the network’s need to efficiently compress features, resulting in fewer neurons than distinct features in the data \cite{elhage2022toy}. To address this, researchers applied dictionary learning and sparse feature extraction on single-layer toy models to decompose neuron activations into more interpretable units. Sparse representations offer a promising solution to polysemanticity and superposition, with sparse autoencoders reducing neuron overlap to create more monosemantic feature representations. \citet{cunningham2023sparse} further explored this approach to extract interpretable features from LLM activations, enabling precise identification of key factors influencing counterfactual behavior, such as decision-making in indirect object identification tasks.

Another aspect of interpretability involves applying these techniques to specific downstream tasks, such as mathematical reasoning or metaphor detection, where understanding model behavior is critical to improving performance. \citet{doshi2017towards} called for more rigorous interpretability studies in these specific contexts, arguing that the demand for transparency is particularly high when models are used in complex and sensitive domains. In response, our work applies sparse decomposition methods to understand how LLMs handle logical reasoning in mathematics or process ambiguous language in metaphor detection. These applications show the practical benefits of interpretability in improving both model performance and trustworthiness.

\begin{figure*}[ht]
    \centering
    \includegraphics[width=1.0\textwidth]{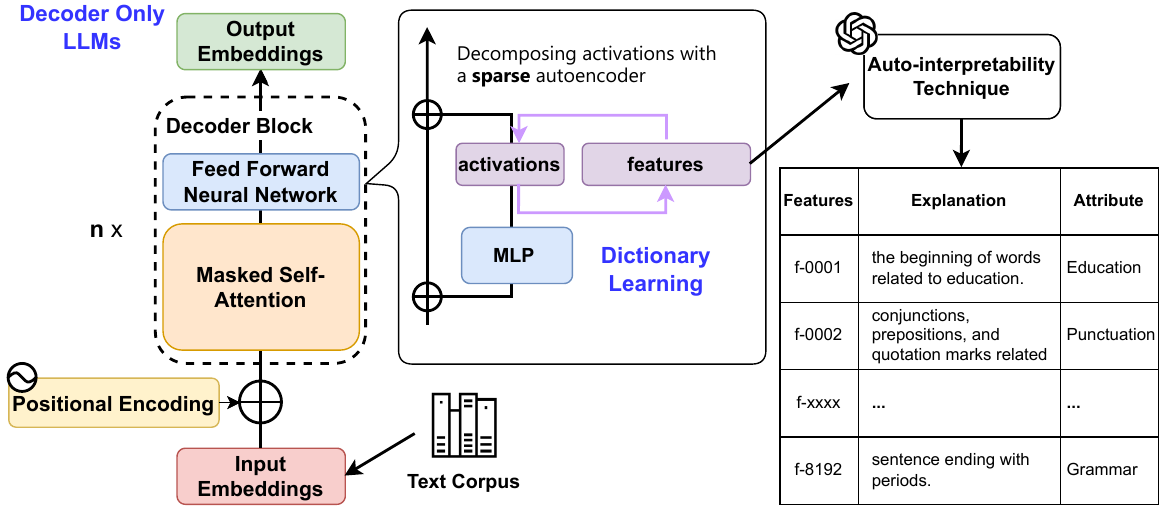}
    \caption{Framework of decomposing LLMs by Dictionary Learning. The framework extracts features from LLMs using a sparse autoencoder to isolate monosemantic features from polysemantic neurons. Once trained, OpenAI's auto-interpretability techniques~\cite{bills2023language} prompt GPT-4 to verbally describe the meaning of these features.}
    \label{fig:framework}
\end{figure*}

\noindent\textbf{Mathematical question answering} has shifted from traditional symbolic systems like Computer Algebra Systems (CAS) to LLMs, which enable understanding and solving math problems without domain-specific programming~\cite{levonian2023retrieval}. Techniques such as chain-of-thought prompting~\cite{wei2022chain} have improved performance on multi-step problems. 
Despite these advancements, LLMs still face challenges with precise calculations, symbolic ambiguities, and maintaining multi-step consistency~\cite{wei2022chain, razeghi2022impact, ahn2024large}. Recent research has integrated external tools such as symbolic reasoning engines and plugged-in solvers \cite{lewkowycz2022solving}. 

\noindent\textbf{Metaphor detection} has improved with pre-trained models like BERT \cite{devlin-etal-2019-bert} and GPT \cite{NEURIPS2020_GPT3}, particularly through fine-tuned models like MelBERT and RoPPT~\cite{choi2021melbert, wang2023metaphor}.  However, challenges like polysemanticity—where single neurons respond to multiple unrelated inputs—persist, making metaphor detection complex. 
Recent work introduces external tools to reduce polysemanticity, like RoPPT~\cite{wang2023metaphor} using parse trees to filter context noise, and FrameBERT~\cite{li2023framebert} leveraging frame-based knowledge for more interpretable metaphor detection.

%% file: latex/3_methodology.tex
\section{Methodology}

Our experimental work is divided into two parts. In the first part, we employ dictionary learning to decompose LLMs and extract relatively monosemantic features, annotating these features using an auto-interpretability technique to produce human-readable dictionaries. In the second part, we use the explicit dictionary obtained from the first step along with a sparse autoencoder to analyze downstream tasks. We designed and implemented an automated pipeline to complete this step.

\subsection{LLM Decomposition}
In this step, we apply a sparse auto-encoder, a dictionary learning algorithm, to generate features from several open-source LLMs that provide a more monosemantic unit of analysis than individual neurons. Our approach is grounded in extensive prior research, particularly in the use of dictionary learning and related techniques for analyzing neural network activations, as well as broader literature on disentanglement \cite{olshausen1997sparse, lee2006efficient, yun2021transformer, elhage2022toy, bricken2023towards, cunningham2023sparse}.

\noindent\textbf{Sparse Autoencoder}~To analyze the superposition phenomenon in LLMs, we employ sparse dictionary learning \cite{olshausen1997sparse, lee2006efficient} by training a single-hidden-layer sparse autoencoder to decompose neuron activations. The encoder extracts sparse features, while the decoder reconstructs the input, ensuring symmetry through tied weights. During training, we balance reconstruction loss (ensuring accurate input reconstruction) and sparsity loss (applying L1 regularization to enhance feature interpretability), enabling the model to learn compact and semantically meaningful feature representations.

\noindent\textbf{Auto-interpretability Technique}
To measure how interpretable the dictionary features are, the automated approach introduced in~\citet{bills2023language} is adopted. In brief, the auto-interpretability procedure samples text where a specific dictionary feature is activated and prompts an LLM to generate a human-readable description of that feature, using this description to predict the feature’s activation on other text samples.

\subsection{Downstream Applications}
We selected mathematical reasoning and metaphor detection as downstream tasks because they test different but crucial capabilities of language models. Mathematical reasoning emphasizes precise logical thinking and structured problem-solving, while metaphor detection challenges the model's ability to understand abstract figurative language and context. These tasks are complementary, with one focusing on logical precision and the other on semantic adaptability, making them important and representative benchmarks for evaluating diverse aspects of language model performance.

\begin{figure*}[ht]
    \centering
    \includegraphics[width=0.9\textwidth]{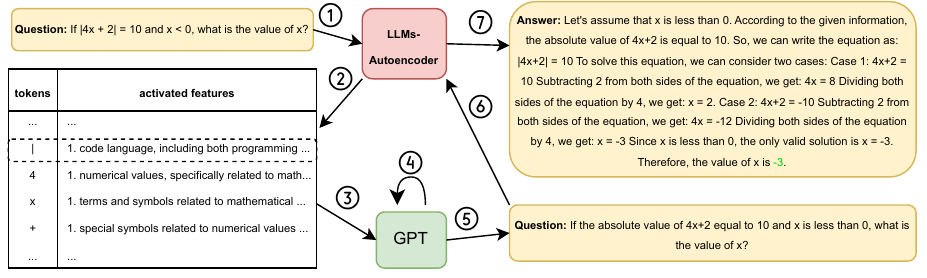}
    \caption{Overview of the pipeline for handling ambiguity in mathematical problem-solving. (1) Activated features are extracted using LLMs with a trained autoencoder plug-in. (2) Rule-based methods analyze token activations to detect and label ambiguous mathematical symbols. (3) GPT-4 is employed to explain these symbols and rephrase the entire question. (4) The original and rephrased questions are compared for equivalence, and if necessary, the question is regenerated. (5) Correctly rephrased question is obtained. (6) The correctly rephrased question is re-input into the LLM. (7) This process results in an enhanced, more accurate answer.}
    \label{fig:framework}
\end{figure*}

\paragraph{Mathematical Reasoning} As seen in \autoref{tab:math_features}, LLMs often fail to prioritize mathematical features when interpreting certain mathematical symbols~\cite{srivatsa2024makes}. In many cases, neither the primary nor secondary activations correspond to features from the mathematical domain. This misalignment introduces ambiguity in processing mathematical problems, leading to potential inaccuracies in understanding and solving such problems. As a result, this issue poses challenges for LLMs in effectively handling tasks involving mathematical reasoning.

Our approach involves a multi-step process to enhance the understanding and answering of ambiguous mathematical questions. 
First, we employ a rule-based method that systematically analyzes the activated features of each token by mapping them to predefined mathematical categories and identifying inconsistencies. This process begins by extracting the top activated features for each mathematical symbol and comparing them against a curated set of features of mathematical attributes. As shown in \autoref{tab:math_features}, if a symbol's dominant features do not align with known mathematical attributes, it is flagged as ambiguous.
Next, GPT-4 is employed to automatically explain these ambiguous symbols within the original problem and rephrase the entire question for clarity.

Once rephrased, the original and modified questions are compared to verify their equivalence by Gemini 1.0-Pro, a language model of comparable capability to GPT-4, ensuring robustness and reducing reliance on a single model. If the two versions are not equivalent, the rephrasing process is repeated to generate a valid rephrased question. Once a correctly rephrased question is obtained, it is then input into the model, resulting in a more accurate and enhanced answer. This process ensures better handling of ambiguity and improved model performance.


\paragraph{Metaphor Detection}~
Our process begins with the LLMs-autoencoder model extracting the activated features of target words from the input, identifying ambiguous targets, and subsequently reformulating the query for clarity. Reformulation is not universally applied but is selectively triggered based on an automated ambiguity detection mechanism. To identify potential semantic ambiguities, we employ GPT-3.5-turbo to analyze these features and determine whether the target word is likely to be misunderstood in the given context. Specifically, GPT-3.5-turbo predicts 
whether the LLM is likely to misinterpret the word’s meaning and thus deciding whether reformulation is necessary. This process is not a simple input transformation but an adaptive self-interpretability approach that enhances LLMs' understanding of complex linguistic contexts.

In \autoref{tab:math_features}, for example, the word "flowed" in a given sentence exhibits its highest activated feature as "terms related to movement or state change of liquid", indicating a literal interpretation rather than its intended metaphorical usage. In this context, "flow" is meant to express abundance, aligning with Explanation 2, "phrases relating to social gatherings or celebrations." When ambiguity is detected, GPT-4 provides targeted supplementary information to clarify the correct meaning of the target word rather than rewriting the entire sentence. This ensures minimal intervention in the language model’s decision-making process while effectively guiding it toward the correct semantic understanding.

Regarding GPT-4’s role in reformulation, our approach does not simply rely on it as an input rewriter but leverages it as a knowledge distillation source, providing explicit semantic guidance. Furthermore, GPT-4 is incorporated based on automated decision rules, meaning that its intervention occurs only when the LLMs-autoencoder predicts a high likelihood of misinterpretation. This design ensures that GPT-4’s involvement is both necessary and minimal, rather than being universally applied. Thus, our method does not depend on GPT-4 as the sole source of supplementary information but rather utilizes it as an external knowledge support to enhance LLM performance in high-ambiguity contexts.

\subsection{Replaceability of Reformulation}
The core objective of this study is to explore an adaptive interpretability method based on LLM-internal features, rather than merely relying on a more powerful model for input reformulation. Our approach does not simply modify inputs but instead leverages LLM-internal feature analysis to predict ambiguity and employs external knowledge to enhance interpretability. Fully relying on GPT-4 for reformulation would effectively outsource the input modality to a stronger model, bypassing the opportunity for LLMs to develop their own reasoning and generalization capabilities. This would not only limit the applicability of our method but also make it difficult to scale to computationally constrained environments. Additionally, our framework is designed as a generalizable interpretability enhancement pipeline, ensuring adaptability to a broader LLM ecosystem rather than restricting it to GPT-4 as an auxiliary tool.

Although this study employs GPT-4 for generating supplementary information, the approach is not inherently dependent on GPT-4 as the only option. The reformulation mechanism is fundamentally driven by LLMs-autoencoder internal feature extraction and analysis. GPT-4 is only triggered when LLMs-autoencoder predicts a high probability of ambiguity, providing necessary supplementary information rather than globally rewriting the input. Furthermore, this approach can be extended to other LLMs for knowledge supplementation or adapted using fine-tuned instruction models (e.g., T5, LLaMA) or rule-based methods to achieve similar semantic enhancement, improving scalability and computational efficiency in resource-constrained environments.

\begin{table*}[tb]\small
\centering
\begin{tabular}{{p{4cm}p{2.5cm}p{8cm}}}
\toprule
\textbf{Math Query} & \textbf{Symbol} & \textbf{Explanation} \\
\hline 
If $\left|4x+2\right|=10$ and $x<0$, what is the value of $x$? & $||$ (absolute value) & 1: code language, including both programming and math functions. \textbf{2: numerical and mathematical expressions or symbols.} 3. URLs, hashtags, and alphanumeric characters. 
 \\
 & $<$ (less than) & 1: punctuation, especially commas and hyphenated numbers, and discourse markers in potentially complex syntax structures such as order, sequence and list. 2: : words and phrases related to personal experiences or events. \textbf{3 special symbols and numerical values.}
 \\ \hline
\textbf{Metaphor Query} & \textbf{Target Word} & \textbf{Explanation} \\ \hline 
The champagne flowed at the wedding. Is the target word 'flowed' a metaphorical or literal expression? & flowed & 1: terms related to movement or state change of liquid. \textbf{2: phrases related to social gathering or celebrations}. 3: rhythmic movement of a musical composition.\\
 \bottomrule
\end{tabular}
\caption{Explanation of activated features for mathematical symbols and metaphorical terms in queries. The activated features, which are associated with mathematical attributes or the correct meanings of the metaphorical language, are in \textbf{bold}.}
\label{tab:math_features}
\end{table*}


\begin{table*}[htb]\small
\centering
\begin{tabular}{p{4cm}p{9.5cm}p{0.75cm}}
\toprule
\textbf{Math Query} & \textbf{Response} & \textbf{Label}\\ \hline
If \textcolor{blue}{$\left|4x+2\right|=10$} and \textcolor{blue}{$x<0$}, what is the value of \textcolor{blue}{$x$}?  &  Let's assume that \textcolor{blue}{$x$} is a positive integer. Then, we can write the equation: \textcolor{blue}{$4x + 2 = 10$} Subtracting \textcolor{blue}{2} from both sides, we get: \textcolor{blue}{$4x = 8$} Dividing both sides by \textcolor{blue}{4}, we get: \textcolor{blue}{$x = 2$} Therefore, the value of \textcolor{blue}{$x$} is \textbf{2}. & \textcolor{red}{\ding{55}} \\ \hline
If the absolute value of \textcolor{blue}{4x+2} equal to \textcolor{blue}{10} and \textcolor{blue}{x} is less than \textcolor{blue}{0}, what is the value of \textcolor{blue}{x}? & Let's assume that \textcolor{blue}{x} is less than \textcolor{blue}{0}. According to the given information, the absolute value of \textcolor{blue}{4x+2} is equal to \textcolor{blue}{10}. So, we can write the equation as: \textcolor{blue}{|4x+2| = 10} To solve this equation, we can consider two cases:Case 1: \textcolor{blue}{4x+2 = 10} Subtracting \textcolor{blue}{2} from both sides of the equation, we get: \textcolor{blue}{4x = 8} Dividing both sides of the equation by \textcolor{blue}{4}, we get: \textcolor{blue}{x = 2}. Case 2: \textcolor{blue}{4x+2 = -10} Subtracting \textcolor{blue}{2} from both sides of the equation, we get: \textcolor{blue}{4x = -12} Dividing both sides of the equation by \textcolor{blue}{4}, we get: \textcolor{blue}{x = -3} Since \textcolor{blue}{x} is less than \textcolor{blue}{0}, the only valid solution is \textcolor{blue}{x = -3}. Therefore, the value of \textcolor{blue}{x} is \textbf{-3}. & \textcolor{deepgreen}{\ding{51}} \\ \hline
\textbf{Metaphor Query} & \textbf{Response} & \textbf{Label}\\ \hline
The champagne flowed at the wedding. Is the target word 'flowed' a metaphorical or literal expression? & The phrase "The champagne flowed at the wedding" is a \textbf{literal} expression. In this context, the word 'flowed' describes the actual movement of champagne being poured and spilling freely as would be expected of sparkling wine. & \textcolor{red}{\ding{55}} \\ \hline
The champagne flowed at the wedding. 'flowed' implies a free and plentiful availability. Is the target word 'flowed' a metaphorical or literal expression? & In the context of the sentence, "The champagne flowed at the wedding," the term 'flowed' is being used in a \textbf{metaphorical} sense. Literally, 'flowed' means to move smoothly and continuously like water or another liquid. Here, it isn't used to describe the actual movement of champagne but rather the abundance and continuous presence of champagne at the wedding, almost like a flowing water current. & \textcolor{deepgreen}{\ding{51}} \\
\bottomrule
\end{tabular}
\caption{The comparison of the pair of Query \& Response between the original question and the reformed question. The responses are generated by Phi-3, and the correct query and response are enhanced by adopting the LLMs-Autoencoder pipeline.}
\label{tab:math_examples}
\end{table*}

%% file: latex/4_experiments.tex
\section{Experiments}
\subsection{Dataset}
\textbf{Training data}~The training data used for the sparse autoencoder in dictionary learning is proportionally sampled from the open-source LLMs pre-training dataset, \textbf{\textit{RedPajama}},\footnote{Dataset available at \url{https://huggingface.co/datasets/togethercomputer/RedPajama-Data-1T}.} ensuring that the capabilities of models across various aspects are preserved.
To improve the model's understanding of downstream tasks, we also incorporated an open-source mathematics dataset, sampling approximately 200M tokens. \textbf{\textit{OpenMathInstruct}}~\cite{toshniwal2024openmathinstruct} is a math instruction tuning dataset generated using permissively licensed Mixtral-8x7B models.

\textbf{Mathematica test data}~(\textbf{\textit{MATH}})~\cite{hendrycks2021measuring} is an open-source math dataset of 12,500 challenging competition mathematics problems. 
Each problem in MATH has a full step-by-step solution to evaluate the reasoning capabilities of models in the domain of mathematics.

\textbf{Metaphor test data}~The \textbf{\textit{MOH}}~\cite{mohammad-etal-2016-metaphor} dataset is constructed by sampling  sentences from WordNet~\cite{miller-1994-wordnet}. Only a single target verb in each sentence is annotated. The average sentence length is 8 tokens, the shortest of our three datasets. 
\textbf{\textit{TroFi}} \cite{birke2006-trofi} consists of sentences from the 1987-89 Wall Street Journal Corpus~\cite{charniak2000-wsj8789}, with an average length of 28.3 tokens per sentence. 

\subsection{Large Language Models}
We evaluate the effectiveness of our method using four state-of-the-art LLMs, each having around 7 billion parameters: Llama 3 (3.1-8B-Instruct)~\cite{dubey2024llama3herdmodels}, Mistral (7B-Instruct)~\cite{jiang2023mistral}, Gemma (7b-it)~\cite{gemma_2024}, and Phi-3 (Small-8K-Instruct)~\cite{abdin2024phi}. These models serve as the backbone of our experimental analysis, where we apply our feature extraction and metaphor detection enhancement techniques to assess their ability to interpret and resolve ambiguous metaphorical expressions.

\subsection{Data Mixing}

\begin{table}[htb]
\small
\centering
\begin{tabular}{lc}
\toprule
\textbf{Source} & \textbf{Tokens (Billion)}\\ \hline
Common Crawl & 8.75 \\
C4 & 1.75 \\
GitHub & 0.60 \\
ArXiv & 0.30 \\
Books & 0.25 \\
Wikipedia & 0.25 \\
StackExchange & 0.20 \\
OpenMathInstruct & 0.20 \\
\hline
\textbf{Total} & \textbf{12.3}\\
\bottomrule
\end{tabular}
\caption{Overview of the data sources used for training, including web crawls (Common Crawl, C4), code repositories (GitHub), scientific articles (arXiv), encyclopedic content (Wikipedia), question-answering platforms (StackExchange), and a specialized dataset for mathematical instruction (OpenMathInstruct). 
}

\label{tab:data_mix}
\end{table}

To ensure that our model maintains strong versatility and performs well in a range of applications, rather than being over-specialized to a single task, we adopted a data mixing training approach. The distribution of data significantly influences the learned weights in the auto-encoder~\cite{elhage2022toy}, particularly in how different features are extracted. To develop a more comprehensive and task-agnostic feature dictionary, we generated the training data by sampling from different LLM pre-training datasets and incorporated portions of downstream task datasets. This approach ensures the learned dictionary is more versatile and applicable to diverse inputs and tasks. The specific data ratios used in training are shown in \autoref{tab:data_mix}.

%% file: latex/5_results.tex
\section{Results}
\subsection{Mathematics Question-Answering}
We confirmed that applying detection and problem-enhancement methods leads to consistent performance improvements across multiple models and mathematical domains. As shown in \autoref{tab:math_results}, all models experienced significant accuracy gains after using the reformed query method, which integrates ambiguity detection and rephrasing techniques.

The results show significant accuracy improvements, shown in \autoref{tab:math_ttest} of Appendix \ref{Sec:A_ttest_results}, particularly in domains requiring complex reasoning. Llama 3 saw the most substantial increase in counting and probability, while Mistral and Phi made notable gains in prealgebra. Gemma improved in both intermediate algebra and prealgebra. These improvements highlight the effectiveness of rephrasing in enhancing the models' ability to handle diverse mathematical problems more effectively. The results of the paired t-test confirm that the performance improvements observed after applying the reformed query method are statistically significant across all models at $\alpha=0.05$. 
These findings provide strong evidence that the ambiguity detection and prompt rephrasing techniques significantly enhance model performance in mathematical question-answering tasks.
 
In summary, across all models and mathematical domains, the reformed query method consistently yielded higher accuracy, confirming the effectiveness of detection and rephrasing techniques in improving the interpretability and performance of mathematical question answering.

\begin{table*}[htb]\small
\resizebox{\textwidth}{!}{
\centering
\begin{tabular}{lcccccccc}
\toprule
\textbf{Model} & \textbf{Intermediate Algebra} & \textbf{Counting/Probability} & \textbf{Precalculus} &\textbf{Number Theory} & \textbf{Algebra} & \textbf{Prealgebra} & \textbf{Geometry} & \textbf{Total} \\ \hline
Llama 3 Original & 27.6 & 23.9 & 26.7 & 23.1 & 34.0 & 37.5 & 34.9 & 30.6 \\
Llama 3 Enhanced &\textbf{ 40.3} & \textbf{47.3} & \textbf{39.1} &\textbf{ 46.7} & \textbf{53.1} & \textbf{63.0} & \textbf{41.1} & \textbf{48.6}
 \\  \hdashline
Mistral Original & 25.8 & 26.1 & 22.4 & 24.4 & 29.8 & 36.6 & 27.7 & 28.3 \\
Mistral Enhanced & \textbf{36.5} & \textbf{39.9} & \textbf{37.9} & \textbf{36.6} & \textbf{41.5} & \textbf{46.8} & \textbf{38.6} & \textbf{41.2}
 \\ \hdashline
Gemma Original & 29.6 & 23.1 & 23.7 & 27.2 & 27.8 & 35.2 & 27.5 &  27.9 \\
Gemma Enhanced & \textbf{40.9} & \textbf{33.8} & \textbf{35.5} & \textbf{35.5} & \textbf{35.1} & \textbf{46.4} & \textbf{39.1} & \textbf{38.6}\\ \hdashline
Phi-3 Original & 16.9 & 18.1 & 21.3 & 17.3 & 25.2 & 30.1 & 28.7 & 22.9 \\
Phi-3 Enhanced &\textbf{ 27.2} & \textbf{29.3} & \textbf{30.5} & \textbf{26.7} & \textbf{35.0} & \textbf{46.1} & \textbf{31.4} & \textbf{33.2} \\
\bottomrule
\end{tabular}
}
\caption{Comparison of answering accuracy across four models (Llama-3, Mistral, Gemma, and Phi-3) in different mathematical domains, using original and reformed query methods. \textbf{Bold} represents the highest-scoring version for each model and question type. In all cases, our automatically enhanced prompts perform the best, with an average absolute increase of 12.52\% (a relative increase of 47.78\%).}
\label{tab:math_results}
\end{table*}

\subsection{Math Error Analysis}

While analyzing a collection of mathematical errors, we categorized mathematical symbols into three distinct types: (1) \textbf{Functions}: Include functions such as \textit{cos}, \textit{sin}, and \textit{sqrt}, among others.
(2) \textbf{Operators}: Consisting of special characters or algebraic symbols, this category includes symbols like \textit{+}, \textit{-}, or algebraic variables such as \textit{x} and \textit{y}. These symbols are often ambiguous due to their polysemantic nature and different usages outside of mathematical contexts (e.g., the less-than sign "<" or absolute value symbols in~\autoref{tab:math_features}).
(3) \textbf{Numbers}: Referring to common numerical digits. \\
Among these types, numbers trigger the least ambiguity during mathematical computations. However, ambiguity is more prevalent in the other two categories, albeit for different reasons. \textbf{Operators} tend to create confusion because of their ubiquity in non-mathematical contexts. For example, as previously discussed, symbols like the less-than sign and the absolute value symbol can be misinterpreted when applied outside of a purely mathematical framework. Variables like \textit{x} and \textit{y} are often included in this category due to their frequent use across various disciplines.
\textbf{Functions} present a more complex challenge,  sometimes being mistakenly identified as code-related functions rather than mathematical functions, complicating interpretation. Furthermore, issues such as tokenization add to the difficulty. For instance, the LaTeX function \textbackslash dfrac (which denotes fractions in display style) can be broken down into tokens like '\textbackslash', 'd', and 'frac', making it even harder to analyze correctly.

\subsection{Metaphor Detection}
Unlike the ambiguity found in mathematical symbols, linguistic ambiguity has long been a significant focus in NLP. 
Our approach directly extracts internal features from the model, allowing for a more explicit and detailed analysis of how the model understands metaphorical expressions.

The results in \autoref{tab:meta_results} reveal insights into the effectiveness of our method in improving metaphor detection with various LLMs. Notably, models like Llama 3 and Mistral demonstrated substantial gains, with Llama 3 improving by 4.3\% on MOH-X and 4\% on Tro-Fi, and Mistral seeing similar increases across both datasets. These improvements suggest that our approach enhances the models' ability to handle the inherent ambiguity in metaphorical expressions. By addressing this challenge, our method enables more accurate metaphor recognition, showing that it outperforms previous approaches in both generalization and handling linguistic complexity.

Additionally, investigation into the understanding of the LLMs via their activations reveals the potential for ambiguity in existing metaphor detection datasets. For instance, the TroFi \cite{birke2006-trofi} dataset states that in "an aerodynamic shape that reduces \textit{drag} 20 \% to 25 \%" and "Their melanin - producing system keeps the skin constantly \textit{filled} with the dark pigment", the italicized words ("drag" and "filled") are being used in a metaphorical sense, whilst we observe highest activation in features relating to their literal interpretations. In such cases, these examples do not align with lay understanding of what constitutes metaphor (which are usually more literary), with both being very close to their literal interpretations.

\begin{table}[tb]\small
\centering
\begin{tabular}{lcc}
\toprule
\textbf{Method} & \textbf{MOH-X} & \textbf{Tro-Fi} \\ \hline
MelBERT & 79.2 & 62.0 \\
MrBERT & 79.8 & 62.7 \\
RoPPT & 80.1 & 63.3 \\
AdMul & 83.9 & 63.3 \\ \hline

Llama 3 &80.1 &61.5 \\ 
Llama 3 Enhanced &\textbf{84.4} &\textbf{65.5} \\ 
\hdashline
Mistral &78.5 &60.8 \\
Mistral Enhanced &\textbf{83.6} &\textbf{63.1} \\
\hdashline
Gemma &78.9 &60.9 \\
Gemma Enhanced &\textbf{82.9} &\textbf{63.8} \\
\hdashline
Phi-3 &76.4 &59.3 \\
Phi-3 Enhanced &\textbf{80.7} &\textbf{62.5} \\\bottomrule
\end{tabular}
\caption{Comparison of metaphor detection accuracy. We include task-specific models and general-purpose LLMs before and after enhancement using internal feature extraction techniques. \textbf{Bold} represents the highest-scoring version for each model and dataset. In all cases, our automatically enhanced prompts perform the best, with an average absolute increase of 3.76\% (a relative increase of 5.38\%).}

\label{tab:meta_results}
\end{table}

%% file: latex/6_conclusions.tex
\section{Conclusion}

We investigated the extent to which the decomposition of traditionally black-box LLMs can aid in increasing their interpretability and lead to performance improvements in downstream tasks such as mathematical reasoning and metaphor detection. we observe improvements from integrating our dictionary learning approach to extract monosemantic features, allowing the identification of which features are activated by a task, to then rewrite the prompt with additional task information and improve performance. We were further able to identify the key misinterpretations being made by LLMs on these tasks via feature activations, in turn identifying the challenges posed by mathematical notation and ambiguous metaphorical usage.

%% file: latex/a_appendix.tex
\section{Paired T-Test Results for Model Performance Improvements on Math}
\label{Sec:A_ttest_results}

To assess the statistical significance of the performance improvements between the original and enhanced versions of the models, we conducted paired t-tests for each model across all mathematical domains. The paired t-test evaluates whether the means of two related groups are statistically different. In this case, we compare the performance (accuracy) of each model's original version against its enhanced version.

The~\autoref{tab:math_ttest} presents the p-values for the paired t-tests for Llama 3, Mistral, Gemma, and Phi-3 models. Since all the p-values are well below the standard significance level of 0.05, we conclude that the improvements in accuracy between the original and enhanced versions are statistically significant for all models.

\begin{table}[htb]\small
\centering
\begin{tabular}{lc}
\toprule
\textbf{Model} & \textbf{p\_value} \\ \hline
Llama3 & $1.5\times 10^{-4} $ \\
Mistral & $2.35\times 10^{-7} $ \\
Gemma & $4.41\times 10^{-7} $ \\
Phi-3 & $1.2\times 10^{-4} $
 \\\bottomrule
\end{tabular}
\caption{All p-values are far below the threshold of 0.05, confirming that the enhancements across all models led to statistically significant improvements in accuracy for mathematical reasoning tasks.}

\label{tab:math_ttest}
\end{table}

\section{Implementation Details}
Our method is model-agnostic and was tested on LLaMA 3, Mistral, Gemma, and Phi-3, all around 7B parameters. We extract activations from a middle-layer MLP block (typically layer 12 or 13), based on prior interpretability research showing that these layers tend to balance low-level syntactic and high-level semantic features. This choice was empirically validated as it yielded the most semantically coherent dictionary features across models.

To train the sparse autoencoder, we freeze all parameters of the base LLM and collect activation vectors from a specific MLP layer. The autoencoder is a single-layer architecture with tied weights and L1 regularization to encourage sparsity. We use the following training configuration: 
\begin{itemize}
    \item Architecture: One-layer sparse autoencoder with ReLU activation
    \item Hidden size: 8192 units (matching 4 times of the target MLP activations)
    \item Optimizer: Adam
    \item Learning rate: 1e-4
    \item Batch size: 16
    \item Training epochs: 10
    \item Loss function: See Eq.\ref{Eq:loss}
\end{itemize}

\begin{equation}
    \mathcal{L}=\underbrace{\frac{1}{\left|X\right|}\sum_{x\in X}^{}\left\|x-\hat{x}\right\|_{2}^{2}}_{\text{Reconstruction\ loss}} +\underbrace{\lambda\left\|s\right\|_{1}}_{\text{Sparsity\ loss}} 
\label{Eq:loss}
\end{equation}
We trained the autoencoder on 12.3B tokens sampled from the RedPajama dataset and OpenMathInstruct (see Table 3). Feature activations were normalized before being passed into the decoder.

Following the methodology introduced by \citeauthor{bills2023language}, we ensured that each training sample was used exactly once during the training of the sparse autoencoder. Our data loader performs a single pass over the dataset without repeated sampling, which prevents the model from overfitting to frequently seen examples and encourages the discovery of generalizable and robust features. All costs of training step are list below:
\begin{table}[htb]
\centering
\begin{tabular}{lc}
\toprule 
 Hardware & A100  \\ \midrule
 Runtime/epoch & 6h  \\ \midrule
 Memory  & 80G \\
  \bottomrule
\end{tabular}
\caption{Experiment details}  
\label{tabel:experiment}
\end{table}

We evaluated the effectiveness of the autoencoder using: \textbf{ reconstruction loss} (MSE): Indicates how well the autoencoder captures the original activation patterns; \textbf{Sparsity level}: Measured as the average number of non-zero activations per input vector; \textbf{Interpretability accuracy}: Following \citeauthor{bills2023language}, we generate human-readable descriptions of dictionary features and test their generalization in new samples.

In our experiments, the average reconstruction error was 0.02 < 0.05, with sparsity 1\% < 5\%. More than 93\% of the interpretable features generated by GPT-4 were validated in unseen contexts with consistent precision.

%% file: latex/acl_latex.bbl
\begin{thebibliography}{47}
\expandafter\ifx\csname natexlab\endcsname\relax\def\natexlab#1{#1}\fi

\bibitem[{Abdin et~al.(2024)Abdin, Jacobs, Awan, Aneja, Awadallah, Awadalla, Bach, Bahree, Bakhtiari, Behl et~al.}]{abdin2024phi}
Marah Abdin, Sam~Ade Jacobs, Ammar~Ahmad Awan, Jyoti Aneja, Ahmed Awadallah, Hany Awadalla, Nguyen Bach, Amit Bahree, Arash Bakhtiari, Harkirat Behl, et~al. 2024.
\newblock Phi-3 technical report: A highly capable language model locally on your phone.
\newblock \emph{arXiv preprint arXiv:2404.14219}.

\bibitem[{Achiam et~al.(2023)Achiam, Adler, Agarwal, Ahmad et~al.}]{OpenAI_GPT4_2023}
Josh Achiam, Steven Adler, Sandhini Agarwal, Lama Ahmad, et~al. 2023.
\newblock \href {https://arxiv.org/abs/2303.08774} {Gpt-4 technical report}.
\newblock \emph{ArXiv}, abs/2303.08774.

\bibitem[{Ahn et~al.(2024)Ahn, Verma, Lou, Liu, Zhang, and Yin}]{ahn2024large}
Janice Ahn, Rishu Verma, Renze Lou, Di~Liu, Rui Zhang, and Wenpeng Yin. 2024.
\newblock \href {https://aclanthology.org/2024.eacl-srw.17} {Large language models for mathematical reasoning: Progresses and challenges}.
\newblock In \emph{Proceedings of the 18th Conference of the European Chapter of the Association for Computational Linguistics: Student Research Workshop}, pages 225--237, St. Julian{'}s, Malta. Association for Computational Linguistics.

\bibitem[{Al-Nazi and Peng(2024)}]{nazi2024large}
Zabir Al-Nazi and Wei Peng. 2024.
\newblock Large language models in healthcare and medical domain: A review.
\newblock In \emph{Informatics}, volume~11, page~57. MDPI.

\bibitem[{Arora et~al.(2018)Arora, Li, Liang, Ma, and Risteski}]{arora2018linear}
Sanjeev Arora, Yuanzhi Li, Yingyu Liang, Tengyu Ma, and Andrej Risteski. 2018.
\newblock \href {https://doi.org/10.1162/tacl_a_00034} {Linear algebraic structure of word senses, with applications to polysemy}.
\newblock \emph{Transactions of the Association for Computational Linguistics}, 6:483--495.

\bibitem[{Belinkov and Glass(2019)}]{belinkov2019analysis}
Yonatan Belinkov and James Glass. 2019.
\newblock \href {https://doi.org/10.1162/tacl_a_00254} {Analysis methods in neural language processing: A survey}.
\newblock \emph{Transactions of the Association for Computational Linguistics}, 7:49--72.

\bibitem[{Bills et~al.(2023)Bills, Cammarata, Mossing, Tillman, Gao, Goh, Sutskever, Leike, Wu, and Saunders}]{bills2023language}
Steven Bills, Nick Cammarata, Dan Mossing, Henk Tillman, Leo Gao, Gabriel Goh, Ilya Sutskever, Jan Leike, Jeff Wu, and William Saunders. 2023.
\newblock \href {https://openaipublic.blob.core.windows.net/neuron-explainer/paper/index.html} {Language models can explain neurons in language models}.
\newblock \emph{OpenAI Blog}.

\bibitem[{Birke and Sarkar(2006)}]{birke2006-trofi}
Julia Birke and Anoop Sarkar. 2006.
\newblock \href {https://aclanthology.org/E06-1042} {A clustering approach for nearly unsupervised recognition of nonliteral language}.
\newblock In \emph{11th Conference of the {E}uropean Chapter of the Association for Computational Linguistics}, pages 329--336, Trento, Italy. Association for Computational Linguistics.

\bibitem[{Bordt et~al.(2022)Bordt, Finck, Raidl, and von Luxburg}]{post-hoc-legal}
Sebastian Bordt, Mich\`{e}le Finck, Eric Raidl, and Ulrike von Luxburg. 2022.
\newblock \href {https://doi.org/10.1145/3531146.3533153} {Post-hoc explanations fail to achieve their purpose in adversarial contexts}.
\newblock In \emph{Proceedings of the 2022 ACM Conference on Fairness, Accountability, and Transparency}, FAccT '22, page 891–905, New York, NY, USA. Association for Computing Machinery.

\bibitem[{Bricken et~al.(2023)Bricken, Templeton, Batson, Chen, Jermyn, Conerly, Turner, Anil, Denison, Askell et~al.}]{bricken2023towards}
Trenton Bricken, Adly Templeton, Joshua Batson, Brian Chen, Adam Jermyn, Tom Conerly, Nick Turner, Cem Anil, Carson Denison, Amanda Askell, et~al. 2023.
\newblock Towards monosemanticity: Decomposing language models with dictionary learning.
\newblock \emph{Transformer Circuits Thread}, 2.

\bibitem[{Brown et~al.(2020)Brown, Mann, Ryder, Subbiah, Kaplan, Dhariwal, Neelakantan, Shyam, Sastry, Askell, Agarwal, Herbert-Voss, Krueger, Henighan, Child, Ramesh, Ziegler, Wu, Winter, Hesse, Chen, Sigler, Litwin, Gray, Chess, Clark, Berner, McCandlish, Radford, Sutskever, and Amodei}]{NEURIPS2020_GPT3}
Tom Brown, Benjamin Mann, Nick Ryder, Melanie Subbiah, Jared~D Kaplan, Prafulla Dhariwal, Arvind Neelakantan, Pranav Shyam, Girish Sastry, Amanda Askell, Sandhini Agarwal, Ariel Herbert-Voss, Gretchen Krueger, Tom Henighan, Rewon Child, Aditya Ramesh, Daniel Ziegler, Jeffrey Wu, Clemens Winter, Chris Hesse, Mark Chen, Eric Sigler, Mateusz Litwin, Scott Gray, Benjamin Chess, Jack Clark, Christopher Berner, Sam McCandlish, Alec Radford, Ilya Sutskever, and Dario Amodei. 2020.
\newblock \href {https://proceedings.neurips.cc/paper_files/paper/2020/file/1457c0d6bfcb4967418bfb8ac142f64a-Paper.pdf} {Language models are few-shot learners}.
\newblock In \emph{Advances in Neural Information Processing Systems}, volume~33, pages 1877--1901. Curran Associates, Inc.

\bibitem[{Charniak et~al.(2000)Charniak, Blaheta, Ge, Hall, Hale, and Johnson}]{charniak2000-wsj8789}
Eugene Charniak, Don Blaheta, Niyu Ge, Keith Hall, John Hale, and Mark Johnson. 2000.
\newblock Bllip 1987-89 wsj corpus release 1.
\newblock \emph{Linguistic Data Consortium, Philadelphia}, 36.

\bibitem[{Choi et~al.(2021)Choi, Lee, Choi, Park, Lee, Lee, and Lee}]{choi2021melbert}
Minjin Choi, Sunkyung Lee, Eunseong Choi, Heesoo Park, Junhyuk Lee, Dongwon Lee, and Jongwuk Lee. 2021.
\newblock \href {https://doi.org/10.18653/v1/2021.naacl-main.141} {{M}el{BERT}: Metaphor detection via contextualized late interaction using metaphorical identification theories}.
\newblock In \emph{Proceedings of the 2021 Conference of the North American Chapter of the Association for Computational Linguistics: Human Language Technologies}, pages 1763--1773, Online. Association for Computational Linguistics.

\bibitem[{Conneau et~al.(2018)Conneau, Kruszewski, Lample, Barrault, and Baroni}]{conneau2018you}
Alexis Conneau, German Kruszewski, Guillaume Lample, Lo{\"\i}c Barrault, and Marco Baroni. 2018.
\newblock \href {https://doi.org/10.18653/v1/P18-1198} {What you can cram into a single {\$}{\&}!{\#}* vector: Probing sentence embeddings for linguistic properties}.
\newblock In \emph{Proceedings of the 56th Annual Meeting of the Association for Computational Linguistics (Volume 1: Long Papers)}, pages 2126--2136, Melbourne, Australia. Association for Computational Linguistics.

\bibitem[{Cunningham et~al.(2023)Cunningham, Ewart, Riggs, Huben, and Sharkey}]{cunningham2023sparse}
Hoagy Cunningham, Aidan Ewart, Logan Riggs, Robert Huben, and Lee Sharkey. 2023.
\newblock Sparse autoencoders find highly interpretable features in language models.
\newblock \emph{arXiv preprint arXiv:2309.08600}.

\bibitem[{Devlin et~al.(2019)Devlin, Chang, Lee, and Toutanova}]{devlin-etal-2019-bert}
Jacob Devlin, Ming-Wei Chang, Kenton Lee, and Kristina Toutanova. 2019.
\newblock \href {https://doi.org/10.18653/v1/N19-1423} {{BERT}: Pre-training of deep bidirectional transformers for language understanding}.
\newblock In \emph{Proceedings of the 2019 Conference of the North {A}merican Chapter of the Association for Computational Linguistics: Human Language Technologies, Volume 1 (Long and Short Papers)}, pages 4171--4186, Minneapolis, Minnesota. Association for Computational Linguistics.

\bibitem[{Doshi-Velez and Kim(2017)}]{doshi2017towards}
Finale Doshi-Velez and Been Kim. 2017.
\newblock Towards a rigorous science of interpretable machine learning.
\newblock \emph{arXiv preprint arXiv:1702.08608}.

\bibitem[{Dubey et~al.(2024)Dubey, Jauhri, Pandey, Kadian, Al-Dahle et~al.}]{dubey2024llama3herdmodels}
Abhimanyu Dubey, Abhinav Jauhri, Abhinav Pandey, Abhishek Kadian, Ahmad Al-Dahle, et~al. 2024.
\newblock \href {http://arxiv.org/abs/2407.21783} {The llama 3 herd of models}.

\bibitem[{Elhage et~al.(2022)Elhage, Hume, Olsson, Schiefer, Henighan, Kravec, Hatfield-Dodds, Lasenby, Drain, Chen et~al.}]{elhage2022toy}
Nelson Elhage, Tristan Hume, Catherine Olsson, Nicholas Schiefer, Tom Henighan, Shauna Kravec, Zac Hatfield-Dodds, Robert Lasenby, Dawn Drain, Carol Chen, et~al. 2022.
\newblock Toy models of superposition.
\newblock \emph{arXiv preprint arXiv:2209.10652}.

\bibitem[{Faruqui et~al.(2015)Faruqui, Tsvetkov, Yogatama, Dyer, and Smith}]{faruqui2015sparse}
Manaal Faruqui, Yulia Tsvetkov, Dani Yogatama, Chris Dyer, and Noah~A. Smith. 2015.
\newblock \href {https://doi.org/10.3115/v1/P15-1144} {Sparse overcomplete word vector representations}.
\newblock In \emph{Proceedings of the 53rd Annual Meeting of the Association for Computational Linguistics and the 7th International Joint Conference on Natural Language Processing (Volume 1: Long Papers)}, pages 1491--1500, Beijing, China. Association for Computational Linguistics.

\bibitem[{Gupta and Sch{\"u}tze(2018)}]{gupta2018lisa}
Pankaj Gupta and Hinrich Sch{\"u}tze. 2018.
\newblock \href {https://doi.org/10.18653/v1/W18-5418} {{LISA}: Explaining recurrent neural network judgments via layer-w{I}se semantic accumulation and example to pattern transformation}.
\newblock In \emph{Proceedings of the 2018 {EMNLP} Workshop {B}lackbox{NLP}: Analyzing and Interpreting Neural Networks for {NLP}}, pages 154--164, Brussels, Belgium. Association for Computational Linguistics.

\bibitem[{Hendrycks et~al.(2021)Hendrycks, Burns, Kadavath, Arora, Basart, Tang, Song, and Steinhardt}]{hendrycks2021measuring}
Dan Hendrycks, Collin Burns, Saurav Kadavath, Akul Arora, Steven Basart, Eric Tang, Dawn Song, and Jacob Steinhardt. 2021.
\newblock \href {https://datasets-benchmarks-proceedings.neurips.cc/paper/2021/file/be83ab3ecd0db773eb2dc1b0a17836a1-Paper-round2.pdf} {Measuring mathematical problem solving with the math dataset}.
\newblock \emph{NeurIPS}.

\bibitem[{Jiang et~al.(2023)Jiang, Sablayrolles, Mensch, Bamford, Chaplot, Casas, Bressand, Lengyel, Lample, Saulnier et~al.}]{jiang2023mistral}
Albert~Q Jiang, Alexandre Sablayrolles, Arthur Mensch, Chris Bamford, Devendra~Singh Chaplot, Diego de~las Casas, Florian Bressand, Gianna Lengyel, Guillaume Lample, Lucile Saulnier, et~al. 2023.
\newblock Mistral 7b.
\newblock \emph{arXiv preprint arXiv:2310.06825}.

\bibitem[{Karpathy et~al.(2015)Karpathy, Johnson, and Fei-Fei}]{karpathy2015visualizing}
Andrej Karpathy, Justin Johnson, and Li~Fei-Fei. 2015.
\newblock Visualizing and understanding recurrent networks.
\newblock \emph{arXiv preprint arXiv:1506.02078}.

\bibitem[{Lee et~al.(2006)Lee, Battle, Raina, and Ng}]{lee2006efficient}
Honglak Lee, Alexis Battle, Rajat Raina, and Andrew Ng. 2006.
\newblock \href {https://proceedings.neurips.cc/paper_files/paper/2006/file/2d71b2ae158c7c5912cc0bbde2bb9d95-Paper.pdf} {Efficient sparse coding algorithms}.
\newblock In \emph{Advances in Neural Information Processing Systems}, volume~19. MIT Press.

\bibitem[{Levonian et~al.(2023)Levonian, Li, Zhu, Gade, Henkel, Postle, and Xing}]{levonian2023retrieval}
Zachary Levonian, Chenglu Li, Wangda Zhu, Anoushka Gade, Owen Henkel, Millie-Ellen Postle, and Wanli Xing. 2023.
\newblock Retrieval-augmented generation to improve math question-answering: Trade-offs between groundedness and human preference.
\newblock \emph{arXiv preprint arXiv:2310.03184}.

\bibitem[{Lewkowycz et~al.(2022)Lewkowycz, Andreassen, Dohan, Dyer, Michalewski, Ramasesh, Slone, Anil, Schlag, Gutman-Solo, Wu, Neyshabur, Gur-Ari, and Misra}]{lewkowycz2022solving}
Aitor Lewkowycz, Anders Andreassen, David Dohan, Ethan Dyer, Henryk Michalewski, Vinay Ramasesh, Ambrose Slone, Cem Anil, Imanol Schlag, Theo Gutman-Solo, Yuhuai Wu, Behnam Neyshabur, Guy Gur-Ari, and Vedant Misra. 2022.
\newblock \href {https://proceedings.neurips.cc/paper_files/paper/2022/file/18abbeef8cfe9203fdf9053c9c4fe191-Paper-Conference.pdf} {Solving quantitative reasoning problems with language models}.
\newblock In \emph{Advances in Neural Information Processing Systems}, volume~35, pages 3843--3857. Curran Associates, Inc.

\bibitem[{Li et~al.(2024)Li, Ai, Chen, Dong, Wu, Liu, Chen, and Tian}]{li2024bladeenhancingblackboxlarge}
Haitao Li, Qingyao Ai, Jia Chen, Qian Dong, Zhijing Wu, Yiqun Liu, Chong Chen, and Qi~Tian. 2024.
\newblock \href {http://arxiv.org/abs/2403.18365} {Blade: Enhancing black-box large language models with small domain-specific models}.

\bibitem[{Li et~al.(2023)Li, Wang, Lin, Guerin, and Barrault}]{li2023framebert}
Yucheng Li, Shun Wang, Chenghua Lin, Frank Guerin, and Lo{\"\i}c Barrault. 2023.
\newblock Framebert: Conceptual metaphor detection with frame embedding learning.
\newblock \emph{arXiv preprint arXiv:2302.04834}.

\bibitem[{Lin et~al.(2024)Lin, Guan, Zhang, Zhang, Li, and Zhang}]{lin2024towards}
Zichao Lin, Shuyan Guan, Wending Zhang, Huiyan Zhang, Yugang Li, and Huaping Zhang. 2024.
\newblock Towards trustworthy llms: a review on debiasing and dehallucinating in large language models.
\newblock \emph{Artificial Intelligence Review}, 57(9):1--50.

\bibitem[{Lipton(2018)}]{lipton2018mythos}
Zachary~C. Lipton. 2018.
\newblock \href {https://doi.org/10.1145/3233231} {The mythos of model interpretability}.
\newblock \emph{Commun. ACM}, 61(10):36–43.

\bibitem[{Mesnard et~al.(2024)Mesnard, Hardin, Dadashi, Bhupatiraju, Pathak et~al.}]{gemma_2024}
Thomas Mesnard, Cassidy Hardin, Robert Dadashi, Surya Bhupatiraju, Shreya Pathak, et~al. 2024.
\newblock \href {http://arxiv.org/abs/2403.08295} {Gemma: Open models based on gemini research and technology}.

\bibitem[{Miller(1994)}]{miller-1994-wordnet}
George~A. Miller. 1994.
\newblock \href {https://aclanthology.org/H94-1111} {{W}ord{N}et: A lexical database for {E}nglish}.
\newblock In \emph{{H}uman {L}anguage {T}echnology: Proceedings of a Workshop held at {P}lainsboro, {N}ew {J}ersey, {M}arch 8-11, 1994}.

\bibitem[{Mohammad et~al.(2016)Mohammad, Shutova, and Turney}]{mohammad-etal-2016-metaphor}
Saif Mohammad, Ekaterina Shutova, and Peter Turney. 2016.
\newblock \href {https://doi.org/10.18653/v1/S16-2003} {Metaphor as a medium for emotion: An empirical study}.
\newblock In \emph{Proceedings of the Fifth Joint Conference on Lexical and Computational Semantics}, pages 23--33, Berlin, Germany. Association for Computational Linguistics.

\bibitem[{Murdoch et~al.(2018)Murdoch, Liu, and Yu}]{murdoch2018beyond}
W~James Murdoch, Peter~J Liu, and Bin Yu. 2018.
\newblock Beyond word importance: Contextual decomposition to extract interactions from lstms.
\newblock \emph{arXiv preprint arXiv:1801.05453}.

\bibitem[{Olah et~al.(2020)Olah, Cammarata, Schubert, Goh, Petrov, and Carter}]{olah2020zoom}
Chris Olah, Nick Cammarata, Ludwig Schubert, Gabriel Goh, Michael Petrov, and Shan Carter. 2020.
\newblock Zoom in: An introduction to circuits.
\newblock \emph{Distill}, 5(3):e00024--001.

\bibitem[{Olshausen and Field(1997)}]{olshausen1997sparse}
B~A Olshausen and D~J Field. 1997.
\newblock Sparse coding with an overcomplete basis set: a strategy employed by v1?
\newblock \emph{Vision Res}, 37(23):3311--3325.

\bibitem[{Razeghi et~al.(2022)Razeghi, Logan~IV, Gardner, and Singh}]{razeghi2022impact}
Yasaman Razeghi, Robert~L Logan~IV, Matt Gardner, and Sameer Singh. 2022.
\newblock \href {https://doi.org/10.18653/v1/2022.findings-emnlp.59} {Impact of pretraining term frequencies on few-shot numerical reasoning}.
\newblock In \emph{Findings of the Association for Computational Linguistics: EMNLP 2022}, pages 840--854, Abu Dhabi, United Arab Emirates. Association for Computational Linguistics.

\bibitem[{Rudin(2019)}]{rudin2019stop}
Cynthia Rudin. 2019.
\newblock Stop explaining black box machine learning models for high stakes decisions and use interpretable models instead.
\newblock \emph{Nature machine intelligence}, 1(5):206--215.

\bibitem[{Shu et~al.(2024)Shu, Zhao, Liu, Demeter, Du, and Zhang}]{shu2024lawllm}
Dong Shu, Haoran Zhao, Xukun Liu, David Demeter, Mengnan Du, and Yongfeng Zhang. 2024.
\newblock Lawllm: Law large language model for the us legal system.
\newblock \emph{arXiv preprint arXiv:2407.21065}.

\bibitem[{Srivatsa and Kochmar(2024)}]{srivatsa2024makes}
KV~Srivatsa and Ekaterina Kochmar. 2024.
\newblock What makes math word problems challenging for llms?
\newblock \emph{arXiv preprint arXiv:2403.11369}.

\bibitem[{Tian et~al.(2024)Tian, Xu, and Mao}]{tian2024theory}
Yuan Tian, Nan Xu, and Wenji Mao. 2024.
\newblock A theory guided scaffolding instruction framework for llm-enabled metaphor reasoning.
\newblock In \emph{Proceedings of the 2024 Conference of the North American Chapter of the Association for Computational Linguistics: Human Language Technologies (Volume 1: Long Papers)}, pages 7731--7748.

\bibitem[{Toshniwal et~al.(2024)Toshniwal, Moshkov, Narenthiran, Gitman, Jia, and Gitman}]{toshniwal2024openmathinstruct}
Shubham Toshniwal, Ivan Moshkov, Sean Narenthiran, Daria Gitman, Fei Jia, and Igor Gitman. 2024.
\newblock Openmathinstruct-1: A 1.8 million math instruction tuning dataset.
\newblock \emph{arXiv preprint arXiv:2402.10176}.

\bibitem[{Wang et~al.(2023)Wang, Li, Lin, Barrault, and Guerin}]{wang2023metaphor}
Shun Wang, Yucheng Li, Chenghua Lin, Loic Barrault, and Frank Guerin. 2023.
\newblock \href {https://doi.org/10.18653/v1/2023.eacl-main.102} {Metaphor detection with effective context denoising}.
\newblock In \emph{Proceedings of the 17th Conference of the European Chapter of the Association for Computational Linguistics}, pages 1404--1409, Dubrovnik, Croatia. Association for Computational Linguistics.

\bibitem[{Wei et~al.(2022)Wei, Wang, Schuurmans, Bosma, Xia, Chi, Le, Zhou et~al.}]{wei2022chain}
Jason Wei, Xuezhi Wang, Dale Schuurmans, Maarten Bosma, Fei Xia, Ed~Chi, Quoc~V Le, Denny Zhou, et~al. 2022.
\newblock Chain-of-thought prompting elicits reasoning in large language models.
\newblock \emph{Advances in neural information processing systems}, 35:24824--24837.

\bibitem[{Xiong et~al.(2024)Xiong, Cai, Cooper, Ge, Papageorgiou, Sifakis, Giannou, Lin, Yang, Agarwal, Chrysos, Oymak, Lee, and Papailiopoulos}]{Xiong_everything_everywhere}
Zheyang Xiong, Ziyang Cai, John Cooper, Albert Ge, Vasilis Papageorgiou, Zack Sifakis, Angeliki Giannou, Ziqian Lin, Liu Yang, Saurabh Agarwal, Grigorios Chrysos, Samet Oymak, Kangwook Lee, and Dimitris Papailiopoulos. 2024.
\newblock \href {https://doi.org/10.48550/arXiv.2410.05603} {Everything everywhere all at once: Llms can in-context learn multiple tasks in superposition}.
\newblock \emph{arXiv}.

\bibitem[{Yun et~al.(2021)Yun, Chen, Olshausen, and LeCun}]{yun2021transformer}
Zeyu Yun, Yubei Chen, Bruno Olshausen, and Yann LeCun. 2021.
\newblock \href {https://doi.org/10.18653/v1/2021.deelio-1.1} {Transformer visualization via dictionary learning: contextualized embedding as a linear superposition of transformer factors}.
\newblock In \emph{Proceedings of Deep Learning Inside Out (DeeLIO): The 2nd Workshop on Knowledge Extraction and Integration for Deep Learning Architectures}, pages 1--10, Online. Association for Computational Linguistics.

\end{thebibliography}
